\documentclass{article}
\usepackage{ijcai19}

\usepackage{times}
\usepackage{soul}
\usepackage{color}
\usepackage{url}
\usepackage[hidelinks]{hyperref}
\usepackage[utf8]{inputenc}
\usepackage[small]{caption}
\usepackage{graphicx}
\usepackage{amsmath}
\usepackage{booktabs}
\usepackage{algorithm}
\usepackage{algorithmic}
\usepackage{amsfonts}
\usepackage{subfig}
\usepackage{bm}
\title{Efficient Project Gradient Descent for Ensemble Adversarial Attack}

\author{
Fanyou Wu\footnote{Corresponding Author.}
\and
Rado Gazo\and
Eva Haviarova\And
Bedrich Benes
\affiliations
Purdue University\\
\emails
\{wu1297, gazo, ehaviar, bbenes\}@purdue.edu
}
\begin{document}

\maketitle

\begin{abstract}

Recent advances show that deep neural networks are not robust to deliberately crafted adversarial examples which many are generated by adding human imperceptible perturbation to clear input. Consider $l_2$ norms attacks, Project Gradient Descent (PGD) and the Carlini and Wagner (C\&W) attacks are the two main methods, where PGD control max perturbation for adversarial examples while C\&W approach treats perturbation as a regularization term optimized it with loss function together. If we carefully set parameters for any individual input, both methods become similar. In general, PGD attacks perform faster but obtains larger perturbation to find adversarial examples than the C\&W when fixing the parameters for all inputs. In this report, we propose an efficient modified PGD method for attacking ensemble models by automatically changing ensemble weights and step size per iteration per input. This method generates smaller perturbation adversarial examples than PGD method while remains efficient as compared to  C\&W method. Our method won the first place in IJCAI19 Targeted Adversarial Attack competition.
\end{abstract}

\section{Introduction}
Deep neural networks are vulnerable to adversarial examples, which are often generated by adding perturbations to the clear input~\cite{szegedy2013intriguing}. Understanding of how to manipulate adversarial examples can improve model robustness~\cite{arnab2018robustness} and help to develop better training algorithms~\cite{goodfellow2014explaining,kurakin2016adversarial,tramer2017ensemble}. Recently, several methods~\cite{szegedy2013intriguing,kurakin2016adversarial,xiao2018spatially,carlini2017towards} have been proposed to find such  examples. Generally, these methods can be divided into two types, 1)~maximum-allowable attack and 2)~regularization-based attack. Typically, maximum-allowable attacks achieve faster but larger perturbations than regularization-based attacks. Both types of attacks retain a hyper-parameter called step size or learning rate which controls the maximum-allowance of or rate of convergence. Step size, in most cases, is a fixed number, while Decoupling Direction and Norm  (DDN)~\cite{rony2018decoupling} method changes its step size by either as a weight decay or weight increase depending on the label of the current iteration.

\captionsetup[subfigure]{labelformat=empty}
\begin{figure}
\centering
\subfloat[Women's Jacket]{
  \includegraphics[width=23mm]{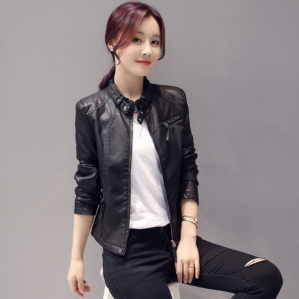}
}
\subfloat[]{
  \includegraphics[width=23mm]{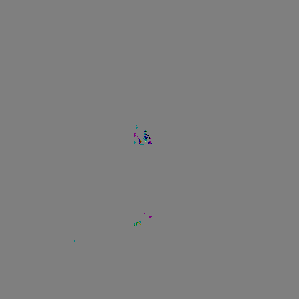}
}
\subfloat[Kid's Jacket]{
  \includegraphics[width=23mm]{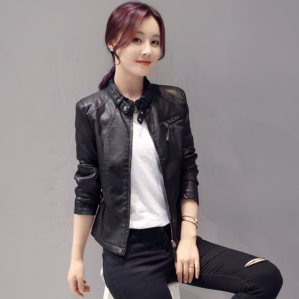}
}
\hspace{0mm}
\subfloat[Kid's Jacket]{
  \includegraphics[width=23mm]{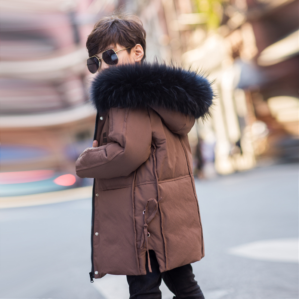}
}
\subfloat[]{
  \includegraphics[width=23mm]{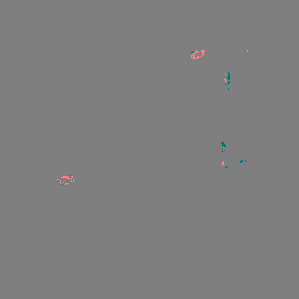}
}
\subfloat[Window Cleaner]{
  \includegraphics[width=23mm]{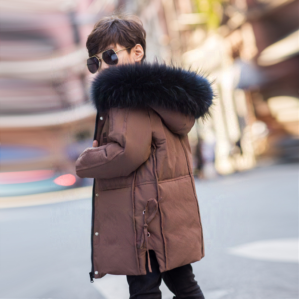}
}
\caption[short]{Examples of adversarial examples generated by ensemble VGG16 and Resnet50~\cite{simonyan2014very,he2016deep}. \textbf{Left column}: the original images. \textbf{Middle column}: the adversarial noises, the gray color represent zero perturbation. \textbf{Right column}: adversarial images generated by our method.}
\end{figure}

Ensemble of models has been commonly used in competitions and researches to enhance performance and improving the robustness~\cite{hansen1990neural,krogh1995neural,caruana2004ensemble}. So attacking ensemble of models that use one set of input is also necessary. Currently, studies  and competition solutions in adversarial attacks are often applied \textit{fuse in logits} with fixed weights as simple ensemble method~\cite{dong2018boosting,rony2018decoupling}.

Inspired by DDN, we propose an efficient project gradient descent for ensemble adversarial attack which improves search direction and step size of ensemble models. Our method won the first place in IJCAI19 Targeted Adversarial Attack competition.  By using the same codes, we ranked twenty-third in Non-Target Track.

\section{Related Work}
We focus on the competition problem formulation and its corresponding metric criteria as well as some well-known attacks.  
\subsection{Problem Formulation}
Let  $\bm x$ be an sample from the input space $\mathcal{X} \in \mathbb{R}^{d\times d \times 3}$ where $d$ is the input width and height, with ground truth label $y^\text{true}$ and random assigned attack target label $y^\text{target}$ from label set $\mathcal{Y}\in \mathbb{N}$. Let $f_i: \mathbb{R}^{d\times d \times 3} \rightarrow \mathbb{R}^M$ be the $i$ attack model with $M$ possbile labels and output logit values. Let $D(\bm x_\text{raw}, \bm x_{\text{adv}})$ be the distance measurement that compare the similarity of clear image $\bm x_\text{raw}$ and adversarial image $\bm x_\text{adv}$  . In this report, we use the following distance measurement which is consistent with competition criteria : 
\begin{equation}D(\bm x_\text{raw},\bm x_\text{adv}|f)=
\left\{
\begin{aligned}
 &64 & failure\\
 &\frac{1}{d^2}\sum_{i=1}^{d} \sum_{j=1}^{d}  ||\bm x^{ij}_\text{raw}-\bm x^{ij}_\text{adv}||_2& success\\
\end{aligned}
\right.
\end{equation} For target attack, success means $\bm x_\text{adv}$ fool the classifier to the assigned label while for non-target attack, success just represent that $\bm x_\text{adv}$ fool the classifier away from its ground truth label $y^\text{true}$. One thing needs to be pointed out:  this distance measurement is not the same as the $l_2$ or so-called Frobenius norm of the $\bm x_\text{raw}-\bm x_\text{adv}$. In this case, it measures the average distance of spatial points in 3 image RGB channels. This specific design encourages us to generate more spatial sparse adversarial examples. \\

Given a set of $N$ evaluation models $f_i$ where $i \in [1,N]$, the final score $S$ is calculated as:   
\begin{align}
    S=\frac{1}{nN}\sum_{i=1}^{N}\sum_{j=1}^{n} D(\bm x^{(j)}_\text{raw},\bm x^{(j)}_\text{adv}|f_i)
\end{align} Where $n$ is the number of testing images. In the final stage, $n$ is equal to 550. For this task, the smaller score, the better the performance.

\subsection{Attacks}
In this section, we review some adversarial examples generated by methods which are related to our methods.

\subsubsection{Fast Gradient Sign Method (FGSM)}
Fast Gradient Sign Method (FGSM) performs a single step update on the original sample x along the direction of the gradient of a loss function $\mathcal{L}(\bm x, y;  \bm \theta)$ ~\cite{szegedy2013intriguing}. The loss could be either an accuracy metric like cross-entropy or a dispersion metric like standard deviation~\cite{jia2019enhancing}.

\begin{align}
    \bm x_\text{adv}=\text{clip}_{[0,1]}\left\{\bm x+\epsilon \cdot \text{sign}(\nabla_{\bm{x}} \mathcal{L}(\bm x, y;  \bm \theta) \right\}
\end{align} where epsilon controls the maximum $l_\infty$ perturbation of the adversarial samples, and the $\text{clip}_{[a,b]}(x)$ function forces x to reside in the range of $[a, b]$. FGSM can be easily extended to $l_2$ norm criteria which fit the competition criteria better. In the following section, we will focus on $l_2$ norms to make this report more related to the competition scenario.

\begin{align}
    \bm x_\text{adv}=\text{clip}_{[0,1]}\left\{\bm x+\epsilon \cdot \frac{\nabla_{\bm{x}} \mathcal{L}(\bm x, y;  \bm \theta)}{||\nabla_{\bm{x}} \mathcal{L}(\bm x, y;  \bm \theta)||_2} \right\}
\end{align}

\vspace{2ex}
\subsubsection{Projected Gradient Descent (PGD)}
Projected Gradient Descent (PGD) is an iterative version of FGSM.  In each iteration, PGD greedily solves the problem of maximizing the loss function:

\begin{align}
    &\bm x^{(t+1)}_\text{adv}=\text{clip}_{[0,1]}\left\{\bm x^{(t)}_\text{adv}+\epsilon \cdot \frac{\nabla_{\bm{x}} \mathcal{L}(\bm x, y;  \bm \theta)}{||\nabla_{\bm{x}} \mathcal{L}(\bm x, y;  \bm \theta)||_2} \right\}\nonumber\\
    & \bm x^{(0)}_\text{adv}=\bm x_\text{raw}
\end{align}Here, clip is a projection function and can be replaced by other functions like \textit{tanh} to force output of each iteration within the effective range. To further use the second order information, Momentum-based method or other optimization methods like RMSprop and Adam can be applied to speed up the iteration process and enhance the transferability. Momentum-based PGD can be formal as below~\cite{zheng2018distributionally,dong2018boosting}.

\begin{align}
    &\bm g^{(t+1)}=\mu \cdot \bm g^{(t)}+\epsilon \cdot \frac{\nabla_{\bm{x}} \mathcal{L}(\bm x, y;  \bm \theta)}{||\nabla_{\bm{x}} \mathcal{L}(\bm x, y;  \bm \theta)||_2} \nonumber\\
     &\bm x^{(t+1)}_\text{adv}=\text{clip}_{[0,1]}\left\{\bm g^{(t)}+\epsilon \cdot \frac{\nabla_{\bm{x}} \mathcal{L}(\bm x, y;  \bm \theta)}{||\nabla_{\bm{x}} \mathcal{L}(\bm x, y;  \bm \theta)||_2} \right\}\\
    & \bm g^{(0)}_\text{adv}=\bm x_\text{raw}\nonumber
\end{align} Where $\mu$ is a parameter to adjust the balance of the current gradient and the historical gradient.

\subsubsection{Decoupled Direction and Norm}

For NeurIPS 18 adversarial attacks competition,~\cite{rony2018decoupling} propose a method which decouples direction and norm for PGD. DDN, generally says, at each iteration, associates its search direction and step size with its current predicted label. This method performs better and can also speed up the search process. 

\subsubsection{Ensemble Adversarial Attack}
In NIPS 17 adversarial attack competition,~\cite{dong2018boosting} reports that \textit{fuse in logits} with cross-entropy and softmax achieve the best performance. Let $f_i(\bm x)$ becomes the logit output of $i$s model. This ensemble method can be formed as follow:

\begin{align}
    &\bm z(\bm x)=\sum_{i=1}^Nw_if_i(\bm x) \nonumber\\
    &\sigma_i(\bm z)=\frac{e^{z_j}}{\sum_{j=1}^K e^{z_j}}  \forall  i \in \mathcal{Y}\\
    &\mathcal{L}(\bm x, y)=\sum_{i=1}^{N} \mathbb{I}\{i=y\}\sigma_i \nonumber
\end{align} where $\mathbb{I}$ is defined as:

\begin{equation}\mathbb{I}\{i=y\}\overset{\text{def}}{=}
\left\{
\begin{aligned}
&1&i=y\\
&0&i\neq y\\
\end{aligned}
\right.
\end{equation} Here $w_i$ is  $i$s model weight. However, in their work, it is a fixed number and does not change during each iteration.

\section{EPGD}
In this section, first we post the Algorithm~\ref{alg:algorithm:1} EPGD, where the Capital E represents Efficient and Ensemble. Then we discuss and give some explanation of this modification. 
\begin{algorithm}[tb]
\caption{EPGD}
\label{alg:algorithm:1}
\textbf{Input}: Input image $\bm x$, target class $y$ \\
\textbf{Input}: Model $f_i(\bm x)$ and $p_i(\bm x, y)$ for $i \in [1,N]$  \\
\textbf{Input}: Mask $\bm m$ \\
\textbf{Input}: Minimal step size $\eta_\text{min}$, maximal step size $\eta_\text{max}$, maximal iteration $K$, confidence level $c$ \\
\textbf{Output}: adversarial example $\bm x_\text{adv}$
\begin{algorithmic}[1] 
\STATE Initialize $\tilde{\bm x}_0=\bm x, w_i=\frac{1}{N} \forall i \in [1,N]$
\FOR{$t\leftarrow 1$ to $K$}
\STATE $\bm z(\bm x)=\sum_{i=1}^Nw_if_i(\bm x) \nonumber$
\STATE    $\sigma_i(\bm z)=\frac{e^{z_j}}{\sum_{j=1}^K e^{z_j}} \forall  i \in \mathcal{Y}$
\STATE$\mathcal{L}(\bm x, y)=\sum_{i=1}^{N} \mathbb{I}\{i=y\}\sigma_i \nonumber$
\STATE $\bm g=\nabla_{\bm x}\mathcal{L}(\bm x, y)$
\STATE $\bm g\leftarrow \bm m\cdot\bm g$
\STATE $p =\text{clip}_{[0,c]}\{ \sum_{i=1}^N w_i p_i(\bm x, y)\}$
\STATE $\eta=\eta_\text{max}-(\eta_\text{max}-\eta_\text{min})\frac{p}{c}$ 
\STATE $\tilde{\bm x}_t=\tilde{\bm x}_{t-1}-\eta \frac{\bm g}{||\bm g||}$
\IF {$\tilde{\bm x}_t$ is adversarial example for all models}
\STATE \textbf{return} $\tilde{\bm x}_t$
\ELSE
\FOR{$i \leftarrow 1$ to $N$}
\IF {$\tilde{\bm x}_t$ is adversarial example for $f_i$}
\STATE $w_i \leftarrow 0$
\ELSE
\STATE $w_i \leftarrow 1$
\ENDIF
\STATE $w_i\leftarrow \frac{ w_i}{\sum_{j=1}^{N}w_j}$
\ENDFOR
\ENDIF
\ENDFOR
\STATE \textbf{return} $\tilde{\bm x}_t$
\end{algorithmic}
\end{algorithm}

\subsection{Changes of Step Size}
Changes of step size is not a new idea to both training deep neural networks and adversarial attacks. DDN~\cite{rony2018decoupling} modify step size as a weight decay per iteration. Here propose our step size modification method, instead of using exponential decay, we perform a truncated linear min-max scale for step size, formally:
\begin{equation}
\begin{aligned}
&p =\text{clip}_{[0,c]}\{ \sum_{i=1}^N w_i p_i(\bm x, y)\}\\
&\eta=\eta_\text{max}-(\eta_\text{max}-\eta_\text{min})\frac{p}{c}\\
\end{aligned}
\end{equation}where $p_i(\bm x, y)$ represent of probability output of label $y$ of model $f_i$ and $c$ the confidence factor (0.5 in this competition) which truncates and probability of ensemble model. If $p$ is above 0.5 then $\eta$  will be $\eta_\text{min}$, otherwise it is  linear proportion to $p$ in the range of   $[\eta_\text{min},\eta_\text{max}]$. Indeed,this method is a greedy estimation of the best step size, and perform better than DDN weight decay in this competition.

\captionsetup[subfigure]{labelformat=empty}
\begin{figure}
\centering
\subfloat[VGG 16]{
  \includegraphics[width=23mm]{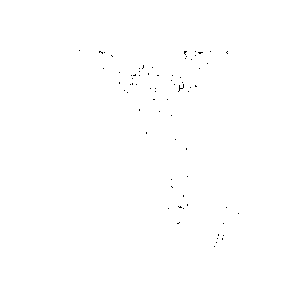}
}
\subfloat[Inception V3]{
  \includegraphics[width=23mm]{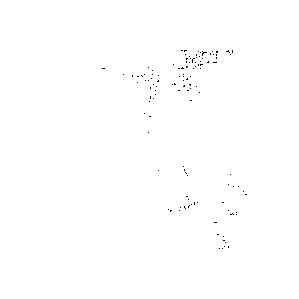}
}
\subfloat[Resnet 50]{
  \includegraphics[width=23mm]{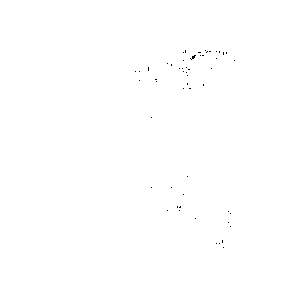}
}
\caption[short]{Examples of adversarial noises generated by single model of VGG 16 Resnet 50 and Inception V3. Here, black points represent that at least one channel value has been modified.}
\end{figure}

\subsection{Changes of Model Ensemble Weights}

Fig.2 shows the general pattern of the individual gradient of input image $\bm x$ with respect to the loss function. In most cases, there are some overlapped regions, which lead to certain transformability for black box attack. We observed that for the fixed model ensemble weights, sometimes a single model has difficulty reaching the decision boundary, which we define as a point when fixed weights lead to the local optima. Different from DDN, other second-order searching methods like the momentum method, could speed up the process. This approach, however, has difficulty reaching the global optima. We address this issue by changing the model ensemble weights. At each iteration, we greedily use the average gradient of all models which were not attacked successfully as the direction to go. Through testing of all the above methods, we found that our method performs the best of all ensemble models.

\subsection{Choices of Masks}
We define mask as $\bm m \in \{0,1\}^{d\times d\times 3}$, and modify the final output as:
\begin{align}
    &\bm x_\text{adv} \leftarrow \bm m \cdot \bm x_\text{adv} +(1-\bm m) \cdot \bm x_\text{raw}
\end{align} Using mask is a very natural idea to $L_2$ attacks, which has been used in both competitions and researches \cite{karmon2018lavan,brendel_brendel_2018}. In our competition criteria, applying spatial masks is better than applying channel masks. For target attack, instead of updating the whole image space, we fixed the noise region as relatively small but continuous area in order reduce overall $l_2$ distance since it is obvious that some features are much more important than others. For non-target attack, disperse grid mask with size $7 \times 7$ and space with 7 pixels performs the best. One possible reason is that most deep neural networks architectures are started with $3 \times 3$ or $7 \times 7$ global pooling operation. \\\\

\section{Implementation and Scores}

\begin{algorithm}[t]
\caption{Final Implementation Algorithm}
\label{alg:algorithm:2}
\textbf{Input}: clear image $\bm x_\text{raw}$, target class $y$, resnet50 model $f_\text{res}$,vgg16 model $f_\text{vgg}$, Inception V3 model $f_\text{inc}$ \\
\textbf{Input}: $T$ number of parameters set $\mathcal{P}_i$ for EPGD, in principle, the smaller index, the smaller noise it will generate \\
\textbf{Output}: adversarial example $\bm x_\text{adv}$
\begin{algorithmic}[1] 
\STATE $ \tilde{\bm{x}}_0\leftarrow \text{EPGD}(\bm x,f_\text{inc})$
\FOR{$t\leftarrow 1$ \textbf{to} $T$}
\STATE $\bm x=\text{EPGD}(\tilde{\bm{x}}_0,f_\text{res},f_\text{vgg},\mathcal{P}_i)$
\IF {$\bm x$ is adversarial example for both $f_\text{res},f_\text{vgg}$}
\STATE \textbf{return} $\bm x$
\ENDIF
\ENDFOR
\STATE \textbf{return} $\bm x$
\end{algorithmic}
\end{algorithm}

\begin{table}[t]
\centering
\begin{tabular}{lll}
\hline
Scenario  & Score \\
\hline
EPGD+Tensorflow+3 Models+All tricks & 34.98     \\
EPGD+Tensorflow+2 Models+All tricks   & 38.62 \\
EPGD+Tensorflow+2 Models+No tricks   & 39.82 \\
PGD+Tensorflow+2 Models+No tricks     & 41.00    \\
PGD+Pytorch+2 Models+No tricks     & 41.64    \\
\hline
\end{tabular}
\caption{Major Millstone for the changes of method and framework}
\label{tab:booktabs}
\end{table}

In a competition, some tricks also help to increase the online score and may have a significant influence on the final rank. Algorithm~\ref{alg:algorithm:2} is our implementation algorithm which addresses some specific issues. Table~\ref{tab:booktabs} is our major milestone for the changes in method and framework.Table~\ref{tab:booktabs:2} is out final submission parameters. In this section, we summarize some useful tricks in adversarial attack competition. For the final submission, we use one  proxy model Inception V3 which was trained by ourselves and two office model which are publicly available Resnet 50 and VGG 16  \cite{simonyan2014very,he2016deep,szegedy2016rethinking}.

\subsection{Float Point Mapping}
Consider the evaluation process which requires us map from $\mathbb{R}^{d \times d \times 3}$ to $\{\mathbb{N}\cap[0,255]\}^{d \times d \times 3}$ as RGB format, the method of mapping float number to integer also affects the attack successful rate as well as the overall distance. We define floor operation as:
\begin{align}
    &\left \lfloor{x}\right \rfloor \overset{\text{def}}{=} \max  \{m\in\mathbb{Z}|M\leq x\}
\end{align} 
Based on this definition, there are three different rounding methods:

\begin{align}
    &\left \lfloor{x_\text{adv}}\right \rfloor \\
    &\left \lfloor{x_\text{adv}+0.5}\right \rfloor \\
    &\lfloor{|x_\text{adv}-x_\text{raw}|}\rfloor \text{sign}(x_\text{adv}-x_\text{raw})+x_\text{raw}
\end{align}Here $x_\text{adv} \in \{\mathbb{R}\cap[0,255]\}$ and $x_\text{raw} \in \{\mathbb{N}\cap[0,255]\}$.

Eq.12 and Eq.13 are the standard definition of \textit{floor} and \textit{round} which no need to discuss much. Eq.14 is our implement of mapping float numbers to integers. The main idea is to round the number as close as possible to raw input. Our method performs the best compared to all three mapping operations in this competition.

\subsection{Resize Methods}

Given the fact that the input size of evaluation models varies while the distance measurement is based on $299\times299$. So the best choice is to generate adversarial examples under $299\times299$. We need a transformation function $T: \mathbb{R}^{299 \times 299 \times 3} \rightarrow \mathbb{R}^{d \times d \times 3}$ where $d$ is required input of the evaluation model. Typically resize transformations are \textit{bilinear interpolating}, \textit{neatest interpolating} and so on. Since the official examples use PIL.Image.BILINEAR as the interpolating method, it becomes very natural to use bilinear interpolating and not surprised it performs best. However, for Tensorflow version no more than  1.13, tf.image.resize\_images does not share the consistent behavior with PIL and will lead to lowering the attack success rate when applying the early-stop trick.\footnote{\url{https://github.com/tensorflow/tensorflow/issues/6720}}.
This problem is issue for tensorflow 1.14 or tensorflow 2.0 given the new v2 image resize api\footnote{\url{https://github.com/tensorflow/tensorflow/releases}}.

\subsection{Parameters Selection}
The ultimate aim for this competition is to generate adversarial example as small as possible within a limited time (550 images in 25 minutes). So using fixed parameters like step size and mask size for each image is not usually optimized: small step size and mask size but larger iteration step will lead to depleting time while large step size and mask size will cause large perturbation. Although the step size modification could partially solve this issue, there should be other tricks to balance it. For the real code implementation, like  Algorithm~\ref{alg:algorithm:2}, we apply a step-like or filter-like strategy, starting from small perturbation parameters and gradually increase the limit of perturbation. \\

\subsection{Adversarial Example Evaluation}

Adversarial example evaluation is a critical step for efficient computing and offline test. Based on the above analysis, we should apply  this step in a cautious manner, since image transformation in different libraries is slightly different. To evaluate exactly the same image for submission, at each iteration for EPGD, we generate a round output $\tilde{\bm x}$ and use PIL.Image.BILINEAR to resize the image as the adversarial example evaluation input.\\

\subsection{Framework Selection}
Tensorflow and Pytorch are the two most important deep learning libraries. In the preliminary stage and half time of the final stage, we use pytorch due to its usability. However, the different behavior of batch-norm implementation between Tensorflow and Pytorch causes us to use Tensorflow since we all know that 2 of 3 given models are used as online evaluation models and run under Tensorflow. 

We also tested several versions of Tensorflow because we know that the permanence and APIs might be altered for different Tensorflow version. It seems that Tensorflow 1.14 could be the best choice for this competition since it has both tf.slim module and resize v2 method. However, we found that the speed is too slow compare to version 1.4. There is another github issue discuss the permanence difference \footnote{\url{https://github.com/tensorflow/tensorflow/issues/25606}}.\\

\begin{table}
\centering
\begin{tabular}{lllllll}
\hline
$T$ & Model&$\eta_\text{min}$&$\eta_\text{max}$& $m_s$ & $K$ & $c$ \\
\hline
0&  $f_\text{inc}$ & 50 & 100 & 50 & 50 & 0.5\\
1&  $f_\text{res},f_\text{vgg}$  & 1 & 100 & 30 & 40 & 0.5\\
2&  $f_\text{res},f_\text{vgg}$  & 1 & 300 & 0 & 40 & 0.5\\
3&  $f_\text{res},f_\text{vgg}$ & 1 & 600 & 0 & 40 & 0.5\\
\hline
\end{tabular}
\caption{Parameter setting for final submission. Here $m_s$ represents the boarder size of mask $\bm m$}
\label{tab:booktabs:2}
\end{table}

\section{Conclusion}
In this report, we propose an efficient modified PGD method called EPGD for attacking ensemble models by automatically changing ensemble weights and step size per iteration, per input. At the same time, we present some useful implementation tools, which aim to search small noise adversarial examples efficiently. Experiments show that our solution can generate smaller perturbation adversarial examples than PGD method, while remaining efficient. With this method, we won the first place in IJCAI19 Targeted Adversarial Attack competition.

\bibliographystyle{named}
\bibliography{ijcai19}

\end{document}